\documentclass{article}
\usepackage{spconf,amsmath,graphicx}
\usepackage{makecell}
\usepackage{multirow}
\usepackage{hyperref}
\usepackage{url}

\title{TRANSFER LEARNING FROM SOUND REPRESENTATIONS \\ FOR ANGER DETECTION IN SPEECH}
\name{Mohamed Ezzeldin A. Elshaer, Scott Wisdom$^*$, Taniya Mishra
\thanks{$^*$S. Wisdom is now with Google.}
\address{
    Affectiva Inc., Boston, MA, USA
    \\
    \{mohamed.ezz, taniya.mishra\}@affectiva.com
}
}
\begin{document}
%
\maketitle
\begin{abstract}
In this work, we train fully convolutional networks to detect anger in speech. Since training these deep architectures requires large amounts of data and the size of emotion datasets is relatively small, we use transfer learning. However, unlike previous approaches that use speech or emotion-based tasks for the source model, we instead use SoundNet, a fully convolutional neural network trained multimodally on a massive video dataset to classify audio, with ground-truth labels provided by vision-based classifiers. As a result of transfer learning from SoundNet, our trained anger detection model improves performance and generalizes well on a variety of acted, elicited, and natural emotional speech datasets. We also test the cross-lingual effectiveness of our model by evaluating our English-trained model on Mandarin Chinese speech emotion data. Furthermore, our proposed system has low latency suitable for real-time applications, only requiring 1.2 seconds of audio to make a reliable classification. 
\end{abstract}

\section{Introduction}
\label{sec:intro}

Owing to the unprecedented success of deep learning in several areas of speech processing such as speech recognition, synthesis, and translation, deep neural networks (DNNs) are now also being used for automatic emotion recognition from speech. A particularly attractive feature of DNNs is that they can consume raw data as input (such as images or audio files) and learn discriminating features from the data as part of the end-to-end optimization process, without requiring human experts to perform feature engineering on the data before the training phase.

However, a significant problem in harnessing the power of deep learning networks for emotion recognition is the mismatch between the large amount of data required by deep networks and the small size of emotion-labeled speech datasets. Training a deep network from scratch requires large quantities of data (on the order of tens of thousands to millions of samples), whereas emotion-labeled datasets are inherently small (on the order of hundreds or thousands of samples) due to the laborious nature of the emotion-labeling task.

In this paper, we focus on the specific emotion recognition task of detecting anger and frustration in speech. This task has many useful applications, including conversational interfaces and social robots, interactive voice response (IVR) systems, market research, customer agent assessment and training, and virtual/augmented reality. To enable these applications to react quickly to user anger and frustration, we enforce the constraint that the model has low latency, only requiring 1.2 seconds of speech audio. Not only does this constraint ensure that the system can be used in near real-time, but 1.2 seconds is also close to the latency of human perception of anger, reported to be 700 milliseconds \cite{rigoulot_feeling_2013}.

For this anger detection task, we adopt an end-to-end convolutional neural network (CNN) that consumes raw audio as its input. To overcome the limited data problem, we use a CNN, SoundNet \cite{aytar_soundnet:_2016}, trained on a large amount of general audio data with ground truth generated by a vision-based teacher model. To train a model that recognizes anger in speech, the weights of our model are initialized with SoundNet and then fine-tuned using a small emotion-labeled dataset, IEMOCAP. We test the generalization performance of this fine-tuned network on both IEMOCAP and various other speech emotion recognition datasets. The IEMOCAP dataset is significantly smaller than the large-scale dataset used to train SoundNet, which consists of two million videos corresponding to over a year of continuous video and audio.

As our main contribution in this paper, we show that effective and low-latency speech emotion recognition models for anger and frustration can be improved by transfer learning from models trained not on a large amount of speech data, but rather on a large dataset of general sounds. We show that despite being fine-tuned on elicited English data, our transfer-learned models generalize well to natural in-the-wild English speech data. Also, our model is effective cross-linguistically on acted Chinese data, albeit with some degradation in performance.

We begin by reviewing required background and related work in section \ref{sec:prior_work}. In section \ref{sec:approach}, we describe our proposed network architecture and training setup. In section \ref{sec:experiments}, we specify the anger detection task, summarize the various datasets used for training and testing, and present and discuss our experimental results. Section \ref{sec:conclusion} provides a conclusion and directions for future work.

\section{Background and relation to prior work}
\label{sec:prior_work}

{\it Transfer learning} is the process of using the knowledge acquired during training a system for one task to train another system for a related task. For many domains, sufficiently large datasets used to train DNNs from scratch are scarce, so transfer learning is often used to solve the problem. 

In the vision modality, Ng et al.\ \cite{ng_deep_2015} used a deep convolutional network initialized with a model pre-trained on ImageNet \cite{russakovsky_imagenet_2015}, then fine-tuned on datasets relevant to facial expressions, followed by fine-tuning on the EmotiW 2015 dataset \cite{dhall_video_2015}. This approach is similar to ours, except we use SoundNet to transfer to a {\it speech} emotion recognition task instead of a visual emotion recognition task. 

For speech emotion recognition, Deng et al.\ \cite{deng_sparse_2013} used sparse autoencoders to learn lower dimensional features across datasets from conventional emotion low-level descriptors (LLDs) that were then used to train a SVM classifier. Gideon et al.\  \cite{gideon_progressive_2017} used progressive neural networks to effectively transfer from speaker and gender classification to emotion recognition across multiple datasets. In contrast to these previous approaches, we perform transfer learning not from a speech or emotion-oriented task, but from a general model of audio provided by SoundNet.

Recently, convolutional neural networks have been shown to be effective for speech emotion recognition, using a variety of conventional input features extracted from raw input audio \cite{khorram_capturing_2017,neumann_attentive_2017, bertero_first_2017}. In fact, several research groups have found that learning end-to-end feature extraction that directly processes raw audio \cite{aytar_soundnet:_2016, trigeorgis_adieu_2016} 
or audio-visual \cite{tzirakis_end--end_2017} 
input can outperform systems that use conventional features. We adopt this approach of learning features for raw audio inputs, and we observe that the features extracted by the end-to-end SoundNet architecture, which is tuned towards modeling general audio events instead of speech, can be effective for anger detection.

The core motivation of our approach 
is that there exist common low-level features between sound and speech. Weninger et al.\ \cite{weninger_acoustics_2013} considered this question, and they found that by selecting various hand-crafted low-level features across the domains of sound, speech, and music, cross-domain learning of continuous arousal and valence measures is feasible. In contrast to this work, we use low-level features learned by a deep network (SoundNet). Also, we consider a binary classification task instead of a regression task. Through our experiments, we reach a similar conclusion: that general sound representations can indeed inform and improve speech emotion recognition.

\section{Approach}
\label{sec:approach}

\begin{figure}[htb]
    \centering
    \centerline{\includegraphics[width=\linewidth]{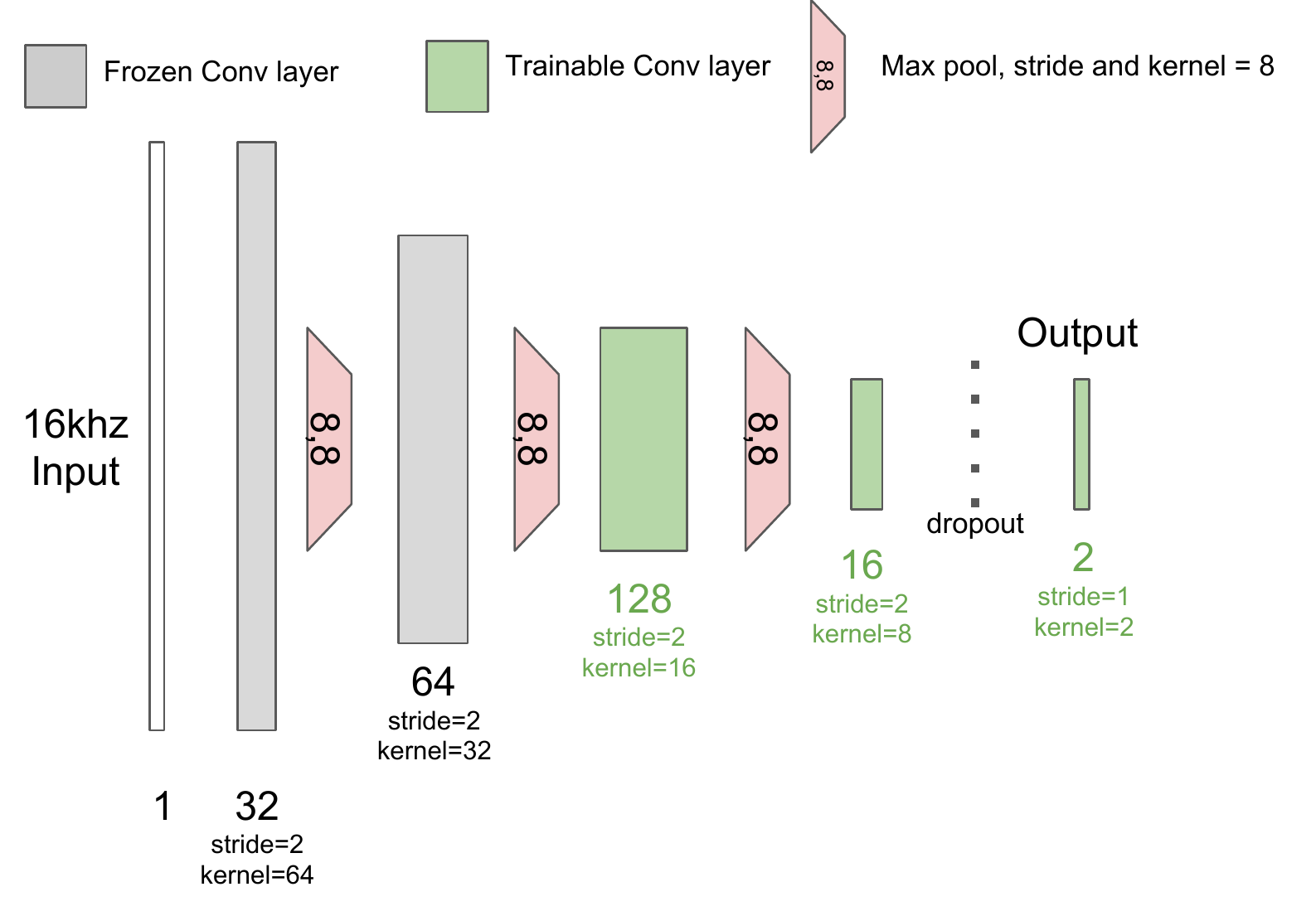}}
    \caption{Network architecture used for transfer learning and learning from scratch. For transfer learning, the frozen layers are colored in gray, and the trainable ones in green. The number of filters, stride, and kernel size are indicated under each convolutional layer.}
\label{fig:arch}
\end{figure}

The deep convolutional network architecture that we use is shown in figure \ref{fig:arch}. The architecture is the same as the 5-layer SoundNet \cite{aytar_soundnet:_2016}. Each one-dimensional convolutional layer is followed by batch normalization and a rectified linear unit (ReLU) activation. We make two modifications to the SoundNet architecture. First, we decrease the number of filters in the fourth layer from 256 to 16, reducing the total number of parameters in the network by a factor of two, from 463,266 to 215,826. We found this change to be essential for avoiding overfitting and making the architecture more suitable for smaller training datasets. Second, we add a dropout layer after the fourth layer as a regularizer.

We train our models on a fixed 1.2 second segments, however because the network is fully convolutional, it accepts variable length inputs. Each segment contains at least 1 second of audio, and segments shorter than 1.2 seconds are zero-padded.


All audio is sampled at 16kHz, despite the SoundNet being originally trained on 22kHz audio. 
Input audio segments are normalized to -20 decibels relative to full-scale (dBFS). This audio is then scaled such that the maximum absolute amplitude is 256, which is similar to the normalization used for training SoundNet \cite{aytar_soundnet:_2016}.

For transfer learning, we load the weights of the first 3 convolutional layers (including the parameters of the corresponding batch normalization) from the first 3 layers of the 5-layer SoundNet. The last 2 layers of our model are initialized randomly. We tried freezing the first 1, 2, or 3 layers of the network during training. Out of these options, we found that freezing the first 2 layers during training yielded the best results on our validation set.

\section{Experiments}
\label{sec:experiments}

In our experiments, the task is a binary classification problem where the goal is to determine if an input 1.2 seconds of speech is angry or not. In this section, we describe the various datasets we use, specify the training procedure for our model, and present and discuss our results.

\subsection{Data}

Table \ref{tab:datasets} reports the amount of data in minutes for each dataset. Notice that the SoundNet training data consists of over a year of audio, which is at least three orders of magnitude larger than any of the emotion datasets. Also, for this anger detection task, these datasets exhibit varying degrees of imbalance in their class labels. For all emotional speech datasets, we use the `anger' and `frustration' tags to indicate positive examples, while other examples are considered negative, except for the `unknown` tags, which were ignored. The rest of this section provides detailed descriptions of these datasets.

\begin{table}[htb]
    \centering
    \begin{tabular}{c|c|c}
         \textbf{Dataset} &
         \multicolumn{2}{c}{\textbf{Amount of data in minutes}} \\
         \hline
         SoundNet training data &
         \multicolumn{2}{c}{$> 525,600$}
         \\
         \hline
         & Anger & Non-Anger \\
         \hline
         IEMOCAP & 227.0 & 343.4 \\
         EmoProsody & 3.6 & 78.0 \\
         {AngerSES} & 195.0 & 54.0 \\
         MASC & 149.0 & 691.0
         
    \end{tabular}
    \caption{Amount of labeled data (in minutes) for each dataset.}
    \label{tab:datasets}
\end{table}

         

\subsubsection{IEMOCAP} 
IEMOCAP \cite{busso_iemocap:_2008} is a corpus of expressive dyadic spoken interactions. It contains approximately 12 hours of audiovisual data including video, speech, and text transcriptions. The data was collected from 10 subjects (5 female and 5 male) and was recorded in 5 sessions. Each session had one male and one female performing improvisations or scripted scenarios designed to elicit particular emotions. The produced utterances were labeled using the three-dimensional valence-activation-dominance scale as well as categorical emotion tags: anger, sadness, happiness, disgust, fear, surprise, frustration, excited and neutral. The corpus consists of over 10000 spoken utterances with an average length of 4.5 seconds. 

\subsubsection{EmoProsody}
The Emotional Prosody Speech and Transcripts (EmoProsody) corpus \cite{mark_liberman_emotional_2002} contains spoken utterances and corresponding transcripts obtained from 8 professional actors (5 females and 3 males) reading a series of semantically neutral utterances in fourteen distinct emotion categories. The corpus consists of 3376 utterances with an average length of 1.5 seconds. 

\subsubsection{AngerSES} 

AngerSES (Anger in Spontaneous English Speech) is a proprietary corpus that was obtained from publicly available sources, with a focus on recordings of spontaneous situations where anger is the most likely emotion displayed. 
Three expert annotators marked the onset and offset of anger and presence of single versus multiple voices in each audio recording in this corpus. This yielded about 4 hours of data that had variations of gender, anger display and intensity, ambient noise and other characteristics. 



\subsubsection{MASC}

The Mandarin Affective Speech Corpus (MASC, \cite{passonneau_masc_2012}) is a Mandarin emotional speech corpus containing elicited utterances from 68 native Mandarin speakers (23 female and 45 male). Subjects read five phrases, fifteen sentences and two paragraphs. Each speaker produced the 5 phrases and 10 sentences three times in five emotional states: neutral, anger, elation, panic and sadness. The paragraphs were produced only in the neutral state. Here, we choose to work with the sentences portion of the data for its appropriate length and availability in all emotions. The sentences portion consists of 20400 utterances, with an average utterance length of 2.2 seconds.

\subsection{Training}

The IEMOCAP dataset is used for training, where sessions 1-3 are used for training, session 4 for validation, and session 5 for testing.
For optimization, we use the Adam algorithm \cite{kingma_adam:_2014} with default parameters except for a learning rate of $5\text{e-}3$. The training loss is cross-entropy between the ground-truth labels and the output of the softmax activation.

The minibatch size is 10. Each minibatch is balanced, consisting of 5 positive and 5 negative examples. 
Each example is a 1.2 second audio clip randomly sampled from an utterance in the training set with a specific label.
Positive examples include segments labeled with anger and frustration, whereas negative examples include segments from any of the other emotions: neutral, happiness, excitement, sadness, fear, surprise.
During training, we augment audio segments using the following random transformations:

\begin{itemize}
    \item Time stretching by a factor uniformly chosen from $[0.9, 1.1]$.
    \item Pitch scaling with number of quarter-steps chosen uniformly from $[-5,5]$.
    \item Adding Gaussian noise with $\mu = 0$ and $\sigma$ chosen uniformly from $[0, 0.005 a]$, where $a$ is the maximum absolute amplitude of any sample in the audio segment.
\end{itemize}

The best model weights are determined by the maximum area under the receiver operating curve (AU-ROC, described in section \ref{sec:results}) measured on the validation set, where the validation set is checked every 200 minibatches. Since positive examples are less numerous than negative examples, the balanced sampling strategy does lead to some oversampling of positive examples during training, but this does not seem to degrade performance.

\subsection{Results}
\label{sec:results}

For our applications, we need to choose a threshold that is applied to the predicted probability of anger to determine the predicted binary label. For example, we may want to be able to set the threshold to achieve a fixed false positive rate. We desire the flexibility to choose this threshold after training, since different applications may require different operating points. To accomplish this flexibility, we use area under the receiver operating curve (AU-ROC) as our main performance metric. The AU-ROC has a statistical interpretation as the probability that a randomly chosen positive example will be ranked higher than a randomly chosen negative example. Furthermore, compared to accuracy, the AU-ROC is more insensitive to class imbalance, which is present in our test sets. 

For evaluation, we extract clips from normalized audio using a sliding window of duration 1.2 seconds and a hop of 600 milliseconds, where we ensure that each 1.2 second clip contains at least one second of audio. Clips that are shorter than 1.2 seconds are zero-padded.

\begin{figure}[htb]%
    \begin{minipage}[b]{.48\linewidth}
      \centering
      \centerline{\includegraphics[width=4.64cm]{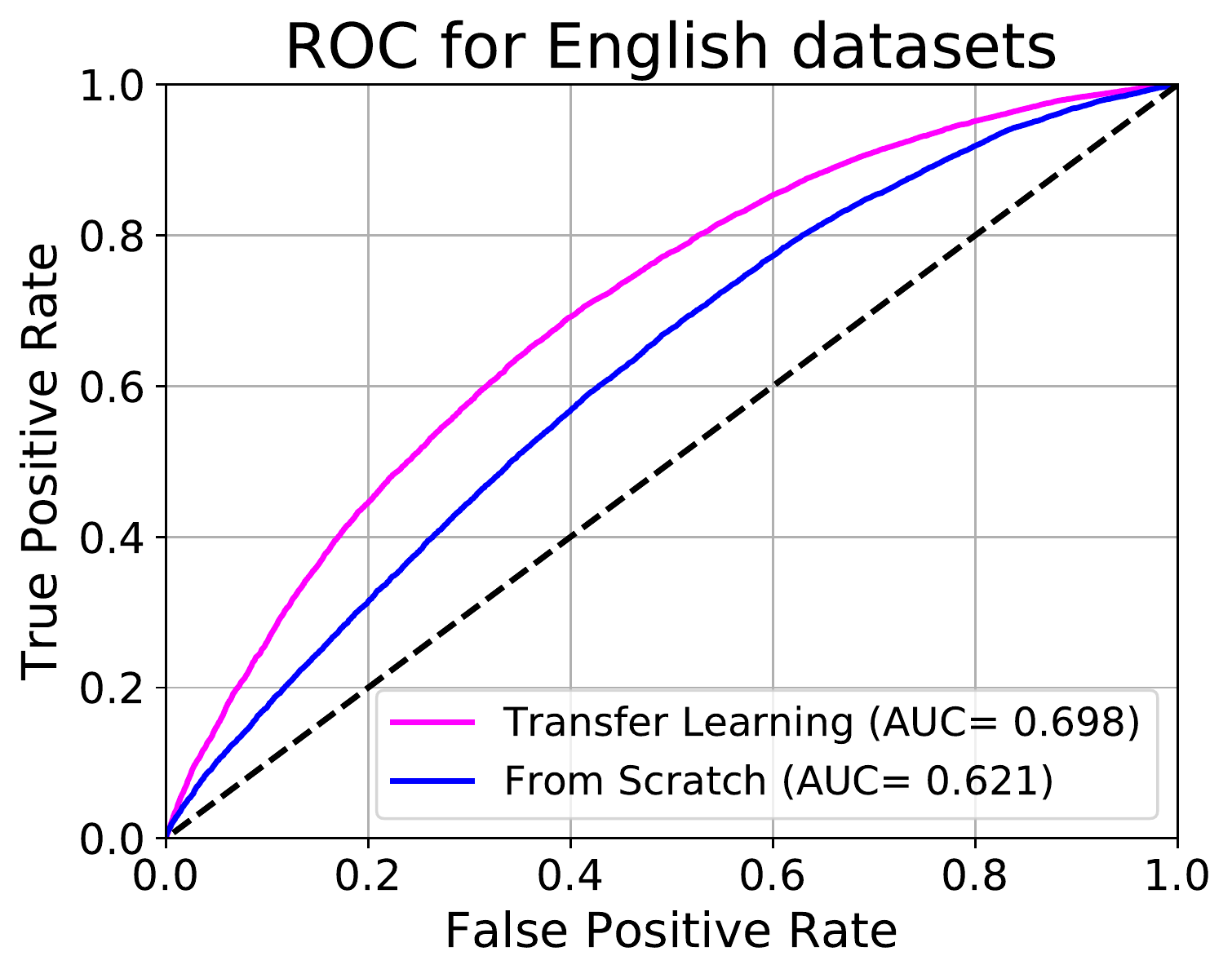}}
    \end{minipage}
    \hfill
    \begin{minipage}[b]{0.48\linewidth}
      \centering
      \centerline{\includegraphics[width=4.64cm]{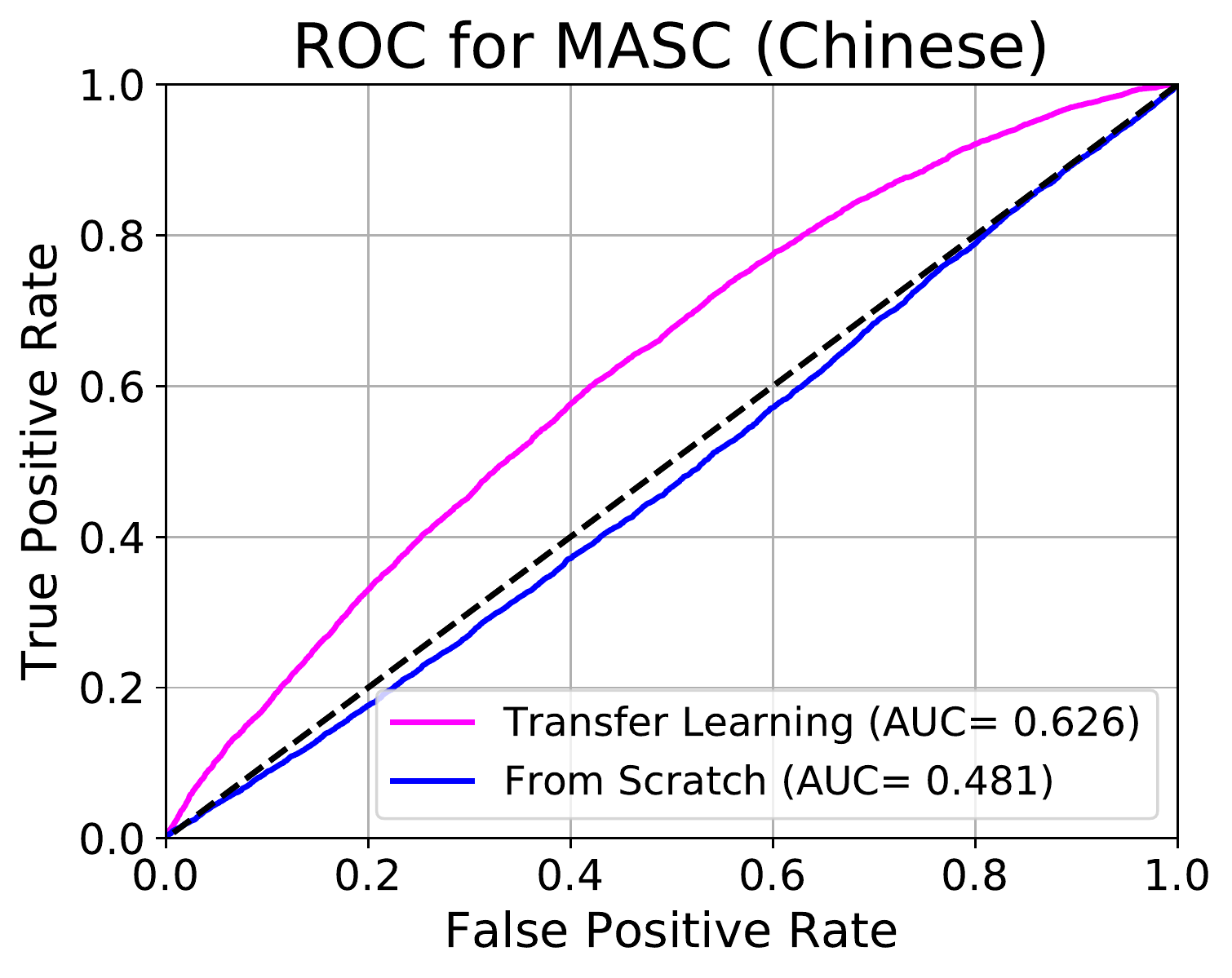}}
    \end{minipage}
    \caption{Receiver operating curve (ROC) for anger detection. The dashed line shows the ROC of a random classifier.}
    \label{fig:rocs}
\end{figure}

The AU-ROC values for the trained-from-scratch model versus transfer-learned model are shown in table \ref{tab:results}. The ROCs for all English data and for Chinese data are shown in figure \ref{fig:rocs}. We tested the statistical significance of the  results by comparing their Mann-Whitney U-statistics using a $z$-test \cite{delong_comparing_1988}. These tests showed that all AU-ROC improvements except the improvement on EmoProsody are significant with $p<3\text{e-}4$ at $99\%$ confidence level.

\begin{table}[h]
    \centering
    \begin{tabular}{c|c|cc}
         Type & Test dataset & Scratch & Transfer  \\
         \hline
         Elicited & IEMOCAP & 0.719 & {\bf 0.736}
         \\
         Acted & EmoProsody & 0.792 & {\bf 0.803}
         \\
         Natural & AngerSES & 0.581 & {\bf 0.669}
         \\
         Mixed & All (English) & 0.621 & {\bf 0.698}
         \\
         \hline
         Acted & MASC (Chinese) & 0.481 & {\bf 0.626}
    \end{tabular}
    \caption{Anger detection results in terms of AU-ROC for various datasets, where ``scratch'' indicates training from random initialization and ``transfer'' indicates transfer from the 5-layer SoundNet with the first two layers frozen. Both models use IEMOCAP for training.}
    \label{tab:results}
\end{table}

\vspace{-15pt}
\subsection{Discussion}

For all datasets used for evaluation, we observed significant improvement in AU-ROC. 
The results in table \ref{tab:results} indicate that the type of data has a significant impact on performance of the model. When trained from scratch on elicited English data (IEMOCAP), the model achieves good performance on elicted English (0.719 AU-ROC on the IEMOCAP test set) and acted English (0.792 AU-ROC on EmoProsody). However, the trained-from-scratch model does not generalize well to natural data, only achieving an AU-ROC of 0.581. Also, the trained-from-scratch model performs poorly on Chinese speech (MASC), achieving an AU-ROC of only 0.481, which is a bit worse than a random classifier.

On the other hand, for the elicted and acted sets, transfer learning improves the AU-ROC a small amount, to 0.736 and 0.803, respectively. But the improvement on both natural English data and acted Chinese data is substantial, with AU-ROCs improving from 0.581 to 0.669 and 0.481 to 0.626, respectively. These results confirm the hypothesis that low-level acoustic representations of natural sounds learned by the hidden layers of SoundNet are general enough to be useful for classifying emotion from speech signals. Also, these acoustic representations improve generalization to unseen conditions including naturalness, ambient noise, and a different language.

\section{Conclusion}
\label{sec:conclusion}


In this paper, we demonstrated that transfer learning from general sound models improves the performance of anger detection in speech.
Using a slightly modified version of the SoundNet architecture, the transfer learning approach outperforms learning from scratch on datasets with different recording conditions and a different language than the training dataset.
This finding suggests that transfer learning from models trained on large amounts of general sound data
are suitable to be fine-tuned for emotion recognition tasks. 

This result is promising because while emotion speech datasets are small and expensive to obtain, 
massive datasets for natural sound events are available, such as the dataset used to train SoundNet \cite{aytar_soundnet:_2016, thomee_yfcc100m:_2016} or Google's AudioSet \cite{gemmeke_audio_2017-1}. These two datasets alone have about 15 thousand hours of labeled audio data. Future work will include further leveraging of this ample data, as well as training models for other speech-based tasks, such as recognizing other types of emotions and affective states.

\vfill\pagebreak

\bibliographystyle{IEEEbib}
\bibliography{icassp2018}

\end{document}